\documentclass[final]{cvpr}

\usepackage{times}
\usepackage{epsfig}
\usepackage{graphicx}
\usepackage{amsmath}
\usepackage{amssymb}

\usepackage{float}

\usepackage[noend]{algpseudocode}
\usepackage{algorithmicx,algorithm}

\usepackage{amsmath}
\usepackage{mathrsfs}

\usepackage{booktabs}
\usepackage{multirow} 

\usepackage{subfigure}

\usepackage{enumitem}
\setenumerate[1]{itemsep=0pt,partopsep=0pt,parsep=\parskip,topsep=5pt}
\setitemize[1]{itemsep=0pt,partopsep=0pt,parsep=\parskip,topsep=5pt}
\setdescription{itemsep=0pt,partopsep=0pt,parsep=\parskip,topsep=5pt}

\usepackage[marginal]{footmisc} 

\usepackage[pagebackref=true,breaklinks=true,colorlinks,bookmarks=false]{hyperref}

\def\cvprPaperID{8040} 
\def\confYear{CVPR 2021}

\begin{document}

\title{Progressive Domain Expansion Network for Single Domain Generalization}

\author{Lei Li$^{12}$, Ke Gao$^{1*}$, Juan Cao$^{1*}$, Ziyao Huang$^{12}$, Yepeng Weng$^{12}$, \\Xiaoyue Mi$^{12}$, Zhengze Yu$^{12}$, Xiaoya Li$^{12}$, Boyang xia$^{12}$\\
$^1$Institute of Computing Technology, Chinese Academy of Sciences, Beijing, China\\
$^2$University of Chinese Academy of Sciences, Beijing, China\\
{\tt\small \{lilei17b,caojuan,huangziyao19f,wengyepeng19s,mixiaoyue19s\}@ict.ac.cn}\\ {\tt\small kegao512@gmail.com, \{yuzhengze,lixiaoya18s,xiaboyang20s\}@ict.ac.cn} 
%
}

\maketitle
\pagestyle{empty}  
\thispagestyle{empty} 
\footnote{*Corresponding author}

\begin{abstract}
	Single domain generalization is a challenging case of model generalization, where the models are trained on a single domain and tested on other unseen domains. A promising solution is to learn cross-domain invariant representations by expanding the coverage of the training domain. These methods have limited generalization performance gains in practical applications due to the lack of appropriate safety and effectiveness constraints. In this paper, we propose a novel learning framework called progressive domain expansion network (PDEN) for single domain generalization. The domain expansion subnetwork and representation learning subnetwork in PDEN mutually benefit from each other by joint learning. For the domain expansion subnetwork, multiple domains are progressively generated in order to simulate various photometric and geometric transforms in unseen domains. A series of strategies are introduced to guarantee the safety and effectiveness of the expanded domains. For the domain invariant representation learning subnetwork, contrastive learning is introduced to learn the domain invariant representation in which each class is well clustered so that a better decision boundary can be learned to improve it's generalization. Extensive experiments on classification and segmentation have shown that PDEN can achieve up to 15.28\% improvement compared with the state-of-the-art single-domain generalization methods. Codes will be released soon at https://github.com/lileicv/PDEN
	
	
\end{abstract}

\section{Introduction}

\begin{figure}[t!]
	\centering
	\subfigure[The traditional decision boundary learned with the original training domain.]{
		\includegraphics[width=0.22\textwidth]{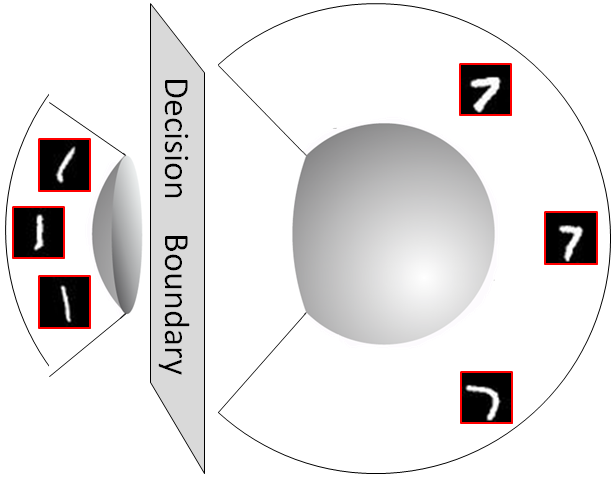}
	}
	\subfigure[The new decision boundary learned with our progressively expanded domains.]{
		\includegraphics[width=0.22\textwidth]{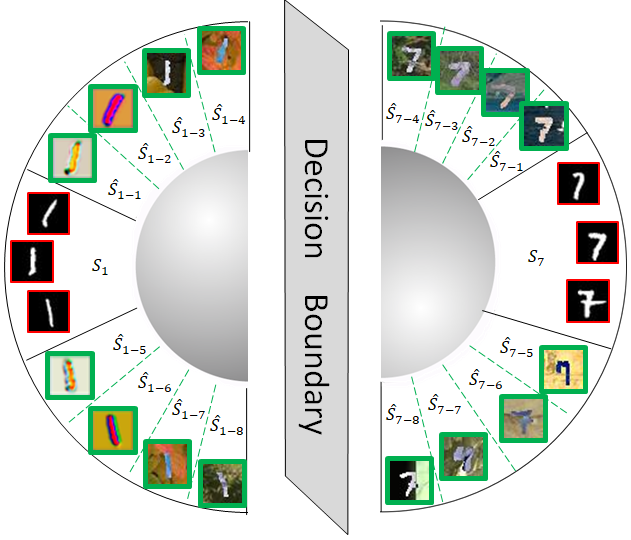}
	}
	\quad
	\subfigure[Domain generalization scenario]{
		\includegraphics[width=0.45\textwidth]{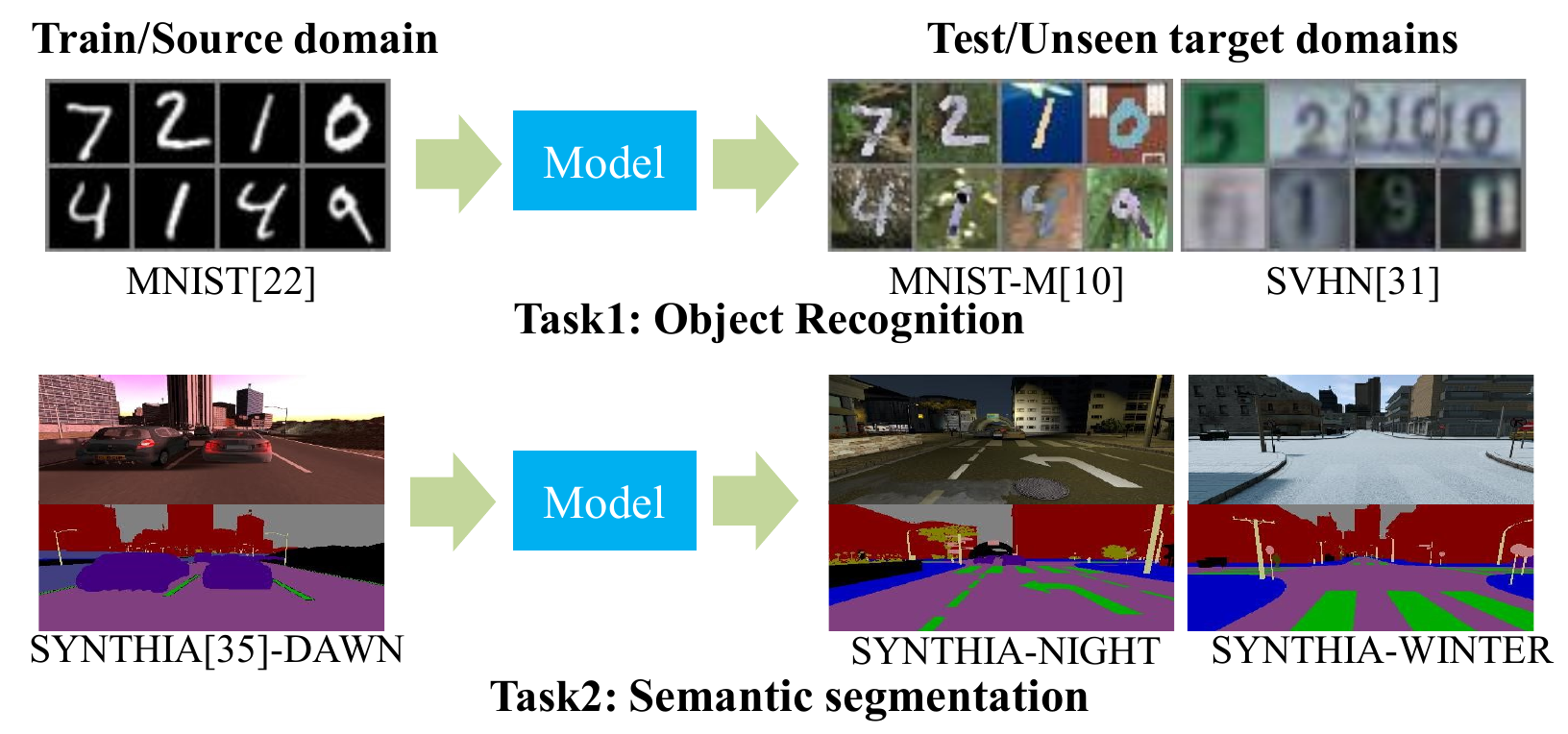}
	}
	\caption{The illustration of our PDEN for single domain generalization. The tiny images in (a) and (b) with red border denote the source domain and the one with green border denote the expanded domains with our PDEN.}
	\label{domain_generalization}
\end{figure}

In this paper, we define domains as various distributions of objects appearance caused by different external conditions(such as weather, background, illumination etc.) or intrinsic attributes(such as color, texture, pose etc.), as shown in Fig.\ref{domain_generalization}. The performance of a deep model usually drops when applied to unseen domains. For example, The accuracy of the CNN model(trained on MNIST) on MNIST test set is 99\%, while that on SVHN test set is only 30\%. Model generalization is important to machine learning.


Two solutions have been proposed to deal with the above issue, namely, domain adaptation \cite{DANN,murez2018image,shu2018a,french2018self} and domain generalization \cite{muandet2013domain,ghifary2015domain,grubinger2017multi,jia2020single}. Domain adaptation aims to generalize to a known target domain whose labels are unknown. Distribution alignment(e.g., MMD) and style transfer(e.g., CycleGAN) are frequently used in these methods to learn domain-invariant features. 
However, it requires data from the target domain to train the model, which is difficult to achieve in many tasks due to lack of data.

Domain generalization, which not requires access to any data from the unseen target domain, can solve these problems. The idea of domain generalization is to learn a domain-agnostic model from one or multiple source domains. Particularity, in many fields we are usually faced with the challenge of giving a single source domain, which is defined as single domain generalization \cite{MADA}. Recently, studies have made progress on this task \cite{2018Train, 2020Generalizing, 2018Beyond, NIPS2018_7779, MADA, zhaoNIPS20maximum}. All of these methods, which are essentially data augmentation, improve the robustness of the model to the unseen domain by extending the distribution of the source domain. Specifically, additional samples are generated by manually selecting the augmentation type\cite{2020Generalizing, 2018Train} or by learning the augmentation through neural networks\cite{MADA, zhaoNIPS20maximum}.

Data augmentation has proved to be an important means for improving model generalization \cite{zhang2016understanding}. However, such methods require the selection of
an augmentation type and magnitude based on the target domain, which is difficult to achieve in other tasks.
They cannot guarantee the safety and effectiveness of synthetic data or even reduce accuracy. \cite{you2015robust, krause20133d}.

In this paper, we propose the progressive domain expansion network (PDEN) to solve the single domain generalization problem. Task models and generators in PDEN mutually benefit from each other through joint learning. Safe and effective domains are generated by the generator under the precise guidance of the task model. The generated domains are progressively expanded to increase the coverage and improve the completeness. Contrastive learning is introduced to learn the cross-domain invariant representation with all the generated domains. It is noteworthy that we can flexibly replace the generator in PDEN to achieve different types of domain expansion.


Our main contributions are as follows:
\begin{itemize}
	\setlength{\itemsep}{0pt}
	\setlength{\parsep}{0pt}
	\setlength{\parskip}{0pt}
	\item We propose a novel framework called progressive domain expansion network (PDEN) for single domain generalization. The PDEN contains domain expansion subnetwork and domain invariant representation learning subnetwork, which mutually benefit from each other by joint learning. 
	
	\item For the domain expansion subnetwork, multiple domains are progressively generated to simulate various photometric and geometric transforms in unseen domains. A series of strategies are introduced to guarantee the safety and effectiveness of these domains.
	
	\item For the domain invariant representation learning subnetwork, contrastive learning is introduced to learn the domain invariant representation in which each class is well clustered so that a better decision boundary can be learned to improve it's generalization.
	
	\item Extensive experiments on classification and segmentation have shown the superior performance of our method. The proposed method can achieve up to 15.28\% improvement compared with other single-domain generalization methods.
\end{itemize}

\begin{figure*}[h]
	\centering
	\subfigure[Progressive domain expansion one by one]{
		\includegraphics[width=0.9\textwidth]{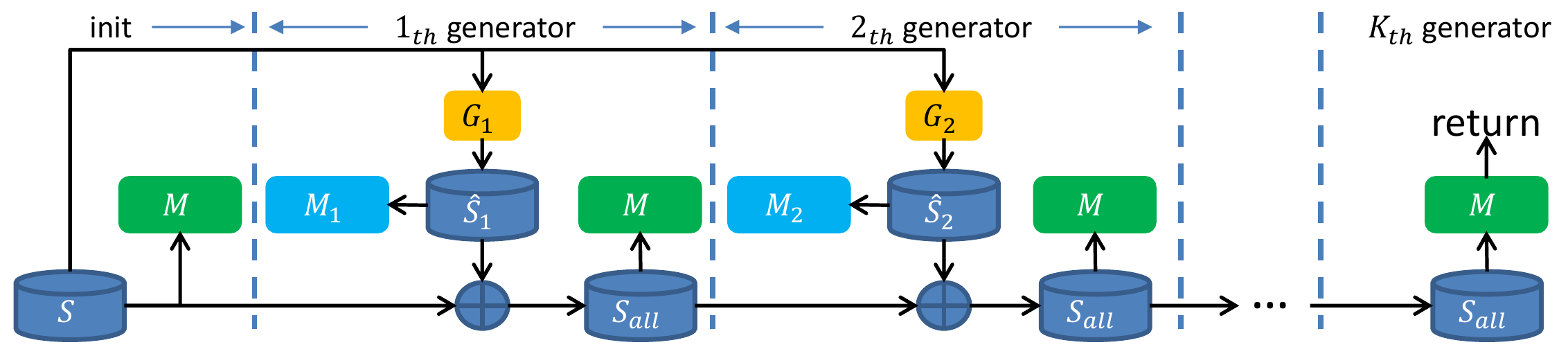}
	}
	\subfigure[Expand the $kth$ unseen domain]{
		\includegraphics[width=0.9\textwidth]{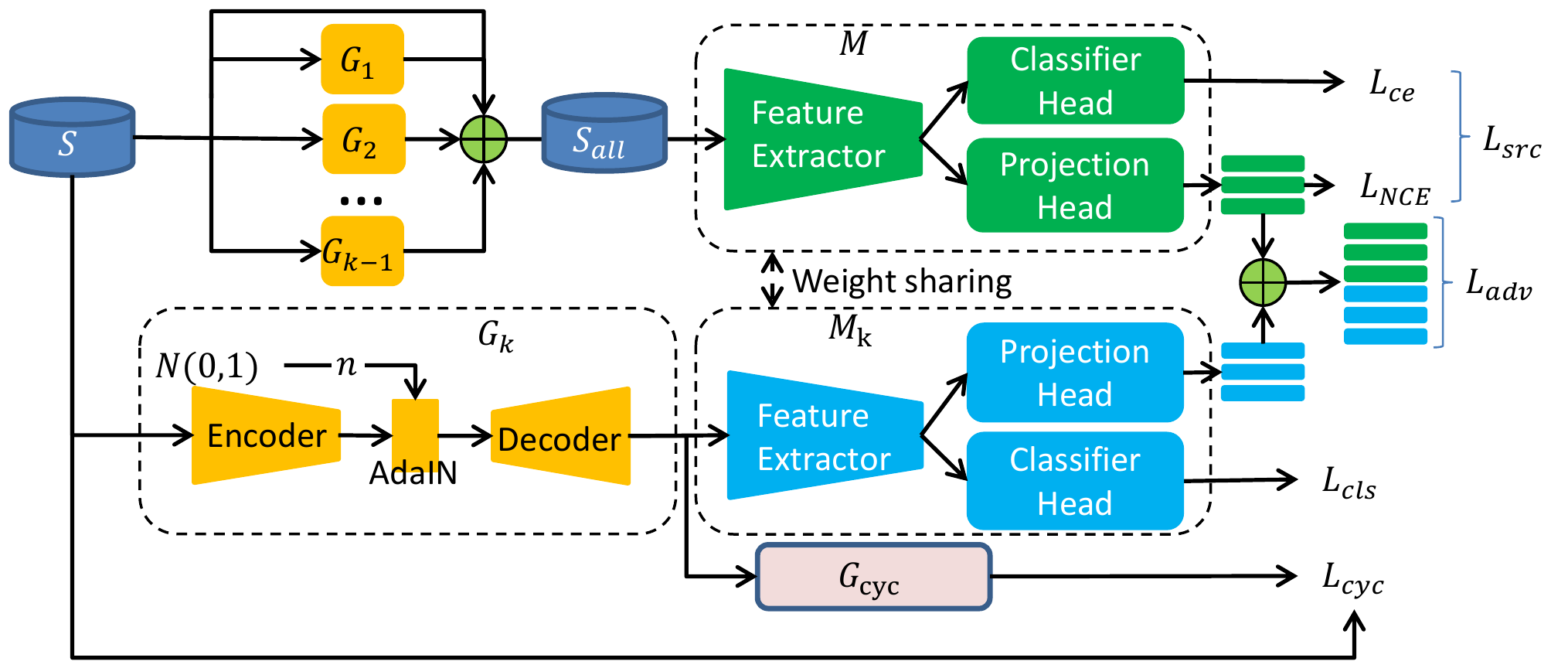}
	}
	\caption{Illustration of the proposed method PDEN. (a) We show how the domain is progressively extended. We trained the task model $M$ and unseen domain generator $G$ alternately. $G$ is trained to synthesize the unseen domain $\hat{\mathcal{S}}$ under the guidance of $M$. Each synthetic domain will be added to the source domain. The task model $M$ will be finetuned after the source domain is updated. (b) We show the network structure of PDEN. Note that $M$ and $M_k$ share the weights, and different $G$ have the same structure but do not share the weights.}
	\label{whole_model}
\end{figure*}

\section{Related Work}

\textbf{Domain Adaptation}. In recent years, many domain adaptation methods \cite{DANN, murez2018image, shu2018a, french2018self} have been proposed to solve the problem of domain drift between source and target domain, including feature-based adaptation\cite{DANN}, instance-based adaptation \cite{dai2007boosting} and model parameter based adaptation \cite{hinton2015distilling}. The domain adaptation method in deep learning is mainly to align the distribution of source domain and target domain, including two kinds of methods: MMD based adaptation method\cite{DDC, DAN} and adversarial based method\cite{DANN}. DDC\cite{DDC} is first proposed to solve the domain adaptation problems in deep networks. DDC fixes the weights of the first 7 layers in AlexNet, and MMD is used on the 8th layer to reduce the distribution difference between the source domain and target domain. DAN\cite{DAN} increased the number of adaptive layers (three in front of the classifier head) and introduced MK-MMD instead of MMD. AdaBN\cite{li2018adaptive} proposed to measure the distribution of the source domain and target domain in BN layer. With the emergence of GAN, a lot of domain adaptation methods based on adversarial learning have been developed. DANN\cite{DANN} is the first research work to reduce the distribution difference between the source domain and target domain by adversarial learning. DSN\cite{DSN} assumes that each domain includes a domain-shared distribution and a domain-specific distribution. Based on this assumption, DSN learned the shared feature and the domain-specific feature respectively. DAAN\cite{DAAN} measures the marginal distribution and conditional distribution with a learnable weight. 

\textbf{Domain Generalization}. Domain generalization is more challenging than domain adaptation.Domain generalization aims to learn the model with data from the source domain and the model can be generalized to unseen domains.

Domain generalization can be categorized as such several research interests: Domain alignment\cite{muandet2013domain, CCSA, li2018domain, dou2019domain} and domain ensemble\cite{mancini2018best}. Domain alignment methods assume that there is a distribution shared by different domains. These methods map the distribution from different domains to the shared one. CCSA\cite{CCSA} propose the contrastive semantic alignment loss to minimize the distance between data with the same label but from different domains and maximize the distance between the data with different class labels. In MMD-AAE\cite{li2018domain}, the feature distribution of source domain and target domain are aligned by MMD, and then the feature representation is matched to the prior Laplace distribution by AAE. Model ensemble\cite{mancini2018best} methods train models for each source domain in the training set, and then ensemble their outputs according to the confidence of each model. 

Single-domain generalization assumes that the training set only contains samples from just one source domain. Recent, many studies have made progress on this task \cite{2018Train, 2020Generalizing, 2018Beyond, NIPS2018_7779, MADA, zhaoNIPS20maximum}. These methods are generally applied to synthesize more samples in image space or feature space to expand the range of data distribution in the training set. BigAug\cite{2020Generalizing} observed that the differences in medical images (such as T2 MRI) are mainly different in 3 aspects: image quality, image appearance, and spatial configuration. They augment more variants for the 3 aspects by data augmentation. However, such methods require the selection of an augmentation type and magnitude based on the target domain, which is difficult to achieve in other tasks. GUD\cite{NIPS2018_7779} and MADA\cite{MADA} synthesize more data through adversarial learning to promote the model's robustness. However, on the one hand, the augmentation type is relatively simple; on the other hand, too much adversarial examples used for training will damage the performance of the classifier.

\textbf{Contrastive Learning}. Contrastive learning is a kind of unsupervised pre-training method for image recognition, which is popular these years. The key idea of contrastive learning is to train a model by bringing the positive pairs closer and pushing apart negative pairs. SimCLR\cite{simclr} generates positive pairs by imposing strong augmentation on whole images. CPC\cite{CPC} utilizes augmentation on image patches and uses the patch-level views for loss optimization. 

\section{Method}

The PDEN proposed in this paper is used for single domain generalization. Suppose the source domain is $\mathcal{S}=\{x_i,y_i\}_{i=1}^{N_S}$, the target domain is $\mathcal{T}=\{x_i, y_i\}_{i=1}^{N_T}$, where $x_i, y_i$ is the $i$-th image and class label, $N_S,N_T$ represent the number of samples in source domain and target domain respectively. The aim is to train the model with only $\mathcal{S}$ then it can be generalized to the unseen $\mathcal{T}$. 


\subsection{The task model $M$}
The overall model architecture of PDEN is shown in Fig. \ref{whole_model} (b), including the task net $M$ and unseen domain generator $G$. In this section, we will introduce the task model in the PDEN.

There are 3 parts in $M$. (1) Feature extractor $F: \mathcal{X}\rightarrow \mathcal{H}$, where $\mathcal{X}$ is the image space and $\mathcal{H}$ is the feature space. $F$ is a stack of convolution layers followed by the pooling layers and activation layers. The output of $F$ is a 1-d vector obtained by global pooling. (2) Classifier head $C: \mathcal{H}\rightarrow \mathcal{Y}$, where $\mathcal{Y}$ is the label space. Here we focus on the classification task, so the task head $C$ is optimized by cross-entropy loss. In our experiment, $C$ is a stack of fully connected layers followed by nonlinear activation layers, and the last activation layer in $C$ is softmax. (3) Projection head $P: \mathcal{H}\rightarrow \mathcal{Z}$, where $\mathcal{Z}$ is the hidden space in which the contrastive loss will be calculated. $P$ contains only one full connection layer in our experiments. We normalize the output vector of $P$ to lie on a unit hypersphere, which enables using an inner product to measure similarity in the $\mathcal{Z}$ space. 

\subsection{The Unseen Domain Generator $G$}

$G$ can convert the original image $x$(original domain $\mathcal{S}$) to a new image $\hat{x}$(unseen domain $\hat{\mathcal{S}}$) as follows:
\begin{equation}
\label{equ_G}
\begin{aligned}
\hat{x} &= G(x, n), n\sim N(0,1)\\
\hat{\mathcal{S}} &=\{(G(x_i, n), y_i)|(x_i, y_i)\in\mathcal{S}\}
\end{aligned}
\end{equation}
where $\hat{x}$ has the same semantic information as $x$, but the domains of $\hat{x}$ and $x$ is different. 

$G$ can be a variety of structures depending on related downstream tasks, such as AutoEncoder \cite{VAE}, HRNet \cite{hrnet}, spatial transform network(STN) \cite{STN} or a combination of these networks. 

\begin{figure}[h]
	\centering
	\subfigure[Domains generated by our domain expansion subnetwork with autoencoder.]{
		\includegraphics[width=0.49\textwidth]{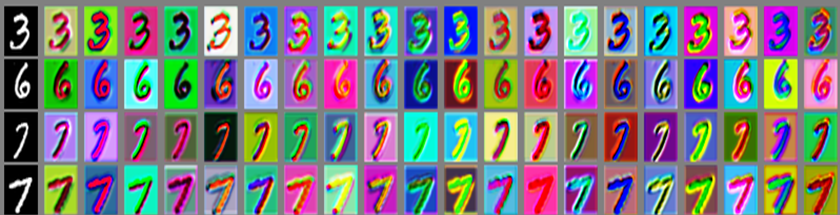}
		\label{gen_by_ae}
	}
	\subfigure[Domains generated by our domain expansion subnetwork with STN.]{
		\includegraphics[width=0.49\textwidth]{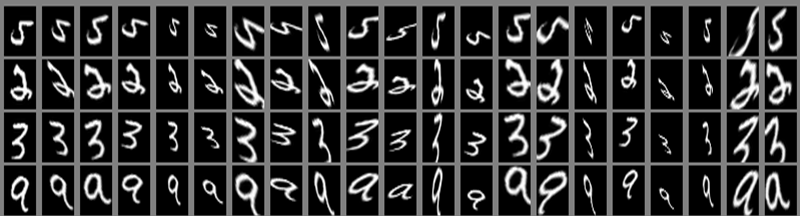}
		\label{gen_by_stn}
	}
	\caption{The domains generated by our domain expansion subnetwork. }
	\label{generated_images}
\end{figure}

\textbf{Autoencoder as $G$:} In our experiment, we mainly use the Autoencoder with AdaIN \cite{styleGAN} as the generator, as the $G_k$ shown in Fig \ref{whole_model}. The generator $G$ contains the encoder $G_E$, the AdaIN and the decoder $G_D$. In AdaIN, there are two fully-connected layers $L_{fc1}, L_{fc2}$:
\begin{equation}
\begin{aligned}
&AdaIN(z, n)=L_{fc1}(n)\frac{z-\mu(z)}{\sigma(z)}+L_{fc2}(n)\\
&G(x, n) = G_D(AdaIN(G_E(x), n))
\end{aligned}
\end{equation}
where $n\sim N(0,1)$. Fig.\ref{gen_by_ae} shows the unseen domains generated by Autoencoder. 

\textbf{STN as $G$:} The Autoencoder can be replaced by the STN\cite{STN} as the generator. The STN is a geometry-aware module which can transform the spatial structure of the image. Fig.\ref{gen_by_stn} shows the unseen domains generated by STN.

PDEN is a framework in which generators can be replaced with different structures depending on the tasks. In our experiment, the autoencoder is applied.

\subsection{Progressive Domain Expansion} 
In order to improve the completeness of the generated domains and expand its coverage, we progressively generate $K$ unseen domains $\{\hat{\mathcal{S}}_k=G_k(\mathcal{S})\}_{k=1}^K$ with the learnable generator $G$.  The task model $M$ is trained with these unseen domains to learn the cross-domain invariant representation. We train the task model and generator alternately, as shown in Fig. \ref{whole_model}. 

Take the $kth$ domain expansion as an example. First, the generator $G$ and task model $M$ are jointly trained to synthesize safe and effective unseen domains $\hat{\mathcal{S}}_k$ by minimize Equ.(\ref{loss_unseen}). Then, the task model $M$ will be retrained with the updated data set $\mathcal{S}\cup\{\hat{\mathcal{S}}_i\}_{i=1}^k$ by minimize Equ.(\ref{loss_src}). The performance of $M$ will be improved, so $M$ can guide the generator $G_{k+1}$ to synthesize better unseen domains. The algorithm is shown in Alg.\ref{alg}.

\subsection{Domain Alignment and Classification}
In this section, we will introduce how to learn cross-domain invariant representation. Given a minibatch $\mathcal{B} = \{x_i, y_i\}_{i=1}^{2N}$, $x_i$ is the source image, $x_i^+=G(x_i,n)$ is the synthetic image originating from $x_i$($x_i$ and $x_i^+$ have the same semantic information, but come from different domains), $y_i$ is the class label. $M$ is optimized by:
\begin{equation}
\begin{aligned}
&L_{ce}(\hat{y}_i, y_i) = \mathop{\rm{min}}_{F,C} -\sum_m y_i^m{\rm log}(\hat{y}_i^m) \\
&L_{NCE}(z_i,z_i^+) = \mathop{\rm{min}}_{F,C} -{\rm log}\frac{exp(z_i\cdot z_i^+)}{\sum_{j=1,j\not=i}^{2N} exp(z_i\cdot z_j)}\\
&L_{src} = L_{ce}(\hat{y}_i, y_i) + L_{NCE}(z_i, z_i^+)
\end{aligned}
\label{loss_src}
\end{equation}
where $y_i^m$ is the $m_{th}$ dimension of $y_i$; $\hat{y}_i=C(F(x_i))$; $z_i=P(F(x_i))$. 

$L_{ce}$ is the cross-entropy loss used for classification. $L_{NCE}$ is the InfoNCE loss\cite{CPC} used for contrastive learning. In the minibatch $\mathcal{B}$, $z_i$ and $z_i^+$ have the same semantic information but come from different domains. By minimizing $L_{NCE}$, the distance between $z_i$ and $z_i^+$ will be smaller. In other words, samples from different domains with the same semantic information will be closer in the $\mathcal{Z}$ space. $L_{NCE}$ will guide $F$ to learn domain-invariant representation.


\subsection{Unseen Domain $\hat{\mathcal{S}}$ Generation}
In this section, we will show how to generate $k$th unseen domain $\hat{\mathcal{S}_k}$ from $\mathcal{S}$ via the generator $G_k$(For convenience, we use $G,\hat{\mathcal{S}}$ instead of $G_k,\hat{\mathcal{S}}_k$). $\hat{\mathcal{S}}$ satisfy the constraints of safety and effectiveness. Safety means the generated samples contain the domain-invariant information. Effectiveness means the generated samples contain various unseen domain-specific information.


\textbf{Safety.} $\hat{\mathcal{S}}$ is safe if all the $x\in\hat{\mathcal{S}}$ can be predicted correctly by task model $M$. Formally, we optimize:
\begin{equation}
\begin{aligned}
L_{cls} = \mathop{min}_{G,F,C}L_{ce}(C(F(G(x,n))), y), n\sim N(0,1)
\end{aligned}
\end{equation}

Cycle consistency loss\cite{cyclegan} is introduced to further ensure the safety of $\hat{\mathcal{S}}$. $\hat{\mathcal{S}}$ is safe if it can be converted to $\mathcal{S}$ by an generator $G_{cyc}$. $G_{cyc}$ has the same structure as $G$, but no noise input.
Formally, we optimize:
\begin{equation}
\begin{aligned}
L_{cyc} = \mathop{min}_{G, G_{cyc}}\Vert x-G_{cyc}(G(x, n))\Vert _2
\end{aligned}
\end{equation}

\textbf{Effectiveness.} Adversarial learning is introduced to generate effective unseen domains. The generator $G$ and task model $M$ are learned jointly. The task model $M$ which extracts the domain-share representation is always trained to minimize the InfoNCE loss. The generator $G$ is trained to maximize the InfoNCE loss. Through adversarial training, $G$ will generate unseen domains from which $M$ can't extract domain shared representation, and $M$ will be better able to extract cross-domain invariant representations. The loss can be defined as:
\begin{equation}
\begin{aligned}
\tilde{L}_{adv}=&\mathop{min}_{G}-L_{NCE}(P(F(x)),P(F(G(x, n)))) + \\
&\mathop{min}_{F,P}L_{NCE}(P(F(x)),P(F(G(x, n))))
\end{aligned}
\label{loss_adv_old}
\end{equation}


However, the loss function Equ. \ref{loss_adv_old} is difficult to converge. As the first item in $\tilde{L}_{adv}$ gets smaller, the gradient gets larger. Therefore, we use the following equation to approximate $\tilde{L}_{adv}$.
\begin{equation}
\begin{aligned}
L_{NCE2}(z_i,z_i^+)&=\sum_i^{2N} {\rm log}\left(1-\frac{exp(z_i\cdot z_i^+)}{\sum_{j=1,j\not=i}^{2N} exp(z_i\cdot z_j)}\right)\\
L_{adv}=&\mathop{min}_{G}-L_{NCE2}(P(F(x)),P(F(G(x, n)))) + \\
&\mathop{min}_{F,P}L_{NCE}(P(F(x)),P(F(G(x, n))))
\end{aligned}
\label{loss_adv}
\end{equation}

We also use a loss function to encourage $G$ to generate more diverse samples.
\begin{equation}
L_{div}= \mathop{min}_G-\Vert G(x, n_1)-G(x, n_2)\Vert_2
\end{equation}
where $n1,n2\sim N(0,1)$, and $n1\not=n2$. To sum up, the loss function of training generate $G$ is as follow:
\begin{equation}
L_{unseen} = L_{cls} + w_{cyc}\cdot L_{cyc} + w_{adv}\cdot L_{adv} + w_{div}\cdot L_{div}
\label{loss_unseen}
\end{equation}

The weight of $L_{cls}$ is always 1, $w_{cyc},w_{adv},w_{div}$ are the weights of $L_{cyc}, L_{adv}, L_{div}$. 
\begin{algorithm}[t]
	\caption{PDEN}
	\label{alg}
	\hspace*{0.02in} {\bf Input:} 
	Source domain dataset $\mathcal{S}$; Pre-train task model $M$; Number of synthetic domains $K$\\
	\hspace*{0.02in} {\bf Output:} 
	learned task model $M$
	\begin{algorithmic}[1]
		\State \textbf{Initialize:} $\mathcal{S}_{all}\leftarrow\mathcal{S}$ 
		\For{k=1,...,K} 
		\State initialize the weights of $G_{k}$ randomly
		\For{t=1,...,$T$} \Comment Train $G_k$ to get $\hat{\mathcal{S}}_k$
		\State Sample $(x_i,y_i)$ from $\mathcal{S}$
		\State $(x_i^+, yi)\leftarrow (G_k(x_i, n), y_i)$
		\State train $G$ and $M$ using Eq.(\ref{loss_unseen})
		\EndFor
		\State Synthetic $kth$ unseen domain $\hat{\mathcal{S}}_k$ using Eq.(\ref{equ_G})
		\State $\mathcal{S}_{all}=\mathcal{S}\cup\hat{\mathcal{S}}_k$
		\For{t=1,...,$T$} \Comment Retrain $M$
		\State Sample $(x_i, y_i)$ from $\mathcal{S}_{all}$
		\State train $M$ using Eq.(\ref{loss_src})
		\EndFor
		\EndFor
		\State \Return $M$
	\end{algorithmic}
\end{algorithm}

\section{Experiment}

\subsection{Datasets and Evaluate}

\begin{table*}[t]
	\begin{center}
		\scalebox{0.90}{
			\begin{tabular}{lcccccc}
				\toprule
				Method & \begin{tabular}[c]{@{}c@{}}Manual Data\\ Augmentation\end{tabular} & SVHN  & MNIST-M& SYNDIGIT	& USPS  & Avg.  \\
				\midrule
				ERM\cite{ERM}  	 &False& 27.83  & 52.72 & 39.65  & 76.94 & 49.29 \\
				CCSA, WVU, 2017\cite{CCSA}   	 &False& 25.89 & 49.29 & 37.31 & 83.72& 49.05\\
				d-SNE, UH, 2019\cite{d-sne}   &False& 26.22 & 50.98 & 37.83 & \textbf{93.16} & 52.05 \\
				JiGen, Huawei, London, 2019\cite{JiGen}    &False& 33.80 & 57.80& 43.79& 77.15& 53.14\\
				GUD, Stanford, 2018\cite{NIPS2018_7779}    &False& 35.51 & 60.41 & 45.32 & 77.26  & 54.62\\
				MADA, UDel, 2020\cite{MADA}  &False& 42.55 & 67.94 & 48.95& 78.53 & 59.49 \\
				PDEN &False& \textbf{62.21}(19.66$\uparrow$)& \textbf{82.20}(14.26$\uparrow$)& \textbf{69.39}(20.44$\uparrow$)& 85.26(6.73$\uparrow$)& \textbf{74.77}(15.28$\uparrow$)  \\
				\midrule
				AutoAugment, Google, 2018 \cite{autoaugment} 		  &True& 45.23 & 60.53 & 64.52  &80.62 & 62.72\\
				RandAugment, Google, 2020 \cite{randaugment}		 &True& 54.77 & 74.05 & 59.60 & 77.33& 66.44\\
				PDEN 		&False& \textbf{62.21}(7.44$\uparrow$)& \textbf{82.20}(8.15$\uparrow$)& \textbf{69.39}( 9.79$\uparrow$)& 85.26(7.93$\uparrow$)& \textbf{74.77}(8.33$\uparrow$) \\
				\bottomrule
			\end{tabular}
		}
	\end{center}
	\caption{Experiment results on Digits dataset. All the models are trained on MNIST. The top half of the table is the comparison with other single domain generalization methods. None of these methods use manual data augmentation. The following section of the table is the comparison with other methods which use manual data augmentation.}
	\label{exp_digits}
\end{table*}

Follow \cite{MADA, NIPS2018_7779}, we evaluated our approach on Digits, CIFAR10-C and SYNTHIA.

\textbf{Digits Dataset}: Digits dataset contains 5 datasets: MNIST\cite{MNIST}, MNSIT-M\cite{DANN}, SVHN\cite{SVHN}, USPS\cite{USPS}, SYNDIGIT\cite{DANN}. Each dataset is considered as a domain. We use MNIST as the source domain and the other four data sets as the target domains. The first 10,000 images in MNIST are used to train the model.  

\textbf{CIFAR10-C Dataset}: We use the CIFAR10\cite{Cifar} as the source domain and the CIFAR10-C\cite{ImageNet-C} as the target domain. CIFAR10-C is a benchmark dataset to evaluate the robustness of classification models. CIFAR10-C dataset consists of test images with 19 corruption types, which are algorithmically generated. The corruptions come from 4 categories and each type of corruption has 5 levels of severity.

\textbf{SYNTHIA Dataset}: The SYNTHIA VIDEO SEQUENCES\cite{SYNTHIA} dataset is used for traffic scene segmentation. The dataset consists of 3 locations: Highway, New York ish and Old European Town. Each location consists of the same traffic situation but under different weather/illumination/season conditions(we use Dawn, Fog, Spring, Night and Winter in our experiment). Following the protocol in\cite{NIPS2018_7779}, we train our model on one domain and evaluate on the other domains. For each domain, we randomly sample 900 images from the left front camera and all the images are resized to $192\times320$ pixels.

\textbf{Evaluate:} For Digit and CIFAR10 datasets, we compute the mean accuracy on each unseen domain. For SYNTHIA dataset, we use the standard mean Intersection over Union(mIoU) to evaluate the performance on each unseen domain.


\subsection{Evaluation of Single Domain Generalization}

We compare our method with the following state-of-the-art methods. (1) Empirical Risk Minimization(ERM) \cite{ERM} is the baseline method trained with only the cross-entropy loss. (2) CCSA \cite{CCSA} aligns samples from different domains of the same category to get the robust feature space for domain generalization. (3) d-SNE\cite{d-sne} minimizes the maximum distance between sample pairs of the same class and maximizes the minimum distance among sample pairs of different categories. (4) GUD \cite{NIPS2018_7779} proposes an adversarial data augmentation method to synthesize more hard samples which can improve the robustness of the classifier. (5) MADA \cite{MADA} minimizes the distance of semantic space and maximize the distance of pixel space to generate more effective samples. (6) JiGen \cite{JiGen} proposes a multi-task learning method that combines the target recognition task and the Jigsaw classification task to improve the cross-domain generalization of the model. (7) AutoAugment(AA) \cite{autoaugment} proposes a method to automatically searches improved data augmentation policies for the specific data set. (8) Based on AA, RandAugment(RA) \cite{randaugment} has a better data augment policies, which greatly reduces the policies space .

\textbf{Comparison on Digits:} We train the model with the first 10,000 images in the MNIST train set, validate the model on the MNIST test set, and evaluate the model on the MNIST-M, SVHN, USPS, and Syndigits datasets.  We calculate the mean accuracy on each data set as the evaluation index. We first compared with the single-domain generalization methods, as shown in the top half of Table \ref{exp_digits}. To be fair, we did not use any manual data augmentation. We observed that our method performs much better than other methods on SVHN, MNIST-M and USPS. On USPS, the performance of our method is comparable to others, mainly because the USPS is more similar to MNIST. The d-SNE\cite{d-sne} performs well on USPS, but bad on other data sets. We also compare with the data augmentation methods as shown in the bottom half of Table \ref{exp_digits}.  The hyperparameters are consistent with those in the original paper. We found that our method performs better than these methods. What's more, our approach is orthogonal to these data augmentation techniques.


\textbf{Comparison on CIFAR10:} We train all the models on the CIFAR10 train set,  validate the models on the CIFAR10 test set, and evaluate the models on the CIFAR10-C. The experimental results across five levels of corruption severity are shown in Tab\ref{exp_cifar10_level}. Our approach performs better than other single-domain generalization methods such as GUD and MADA. The severer  the corruption, the more our approach surpasses MADA. Compared to approaches using manual data augmentation, our approach performs as well as they do at lower corruption levels and outperform them at higher corruption levels. We also show the experimental results across different types of corruptions with the 5th level severity in Tab \ref{exp_cifar10_type}. Our approach has higher average accuracy than other approaches. In some corruption types, the RandAugment approach performs better than us. However, it is important to note that there is no manual data augmentation in our approach, and our approach can be used together with RandAugment.

\begin{table}[t]
	\begin{center}
		\scalebox{0.9}{
			\begin{tabular}{lccccc}
				\hline
				Method                      & Level1 & Level2 & Level3 & Level4 & Level5 \\ \hline
				ERM\cite{ERM}               & 87.8   & 81.5   & 75.5   & 68.2   & 56.1   \\
				GUD\cite{NIPS2018_7779}     & 88.3   & 83.5   & 77.6   & 70.6   & 58.3   \\
				MADA\cite{MADA} & 90.5   & 86.8   & 82.5   & 76.4   & 65.6   \\
				AA\cite{autoaugment}        & 91.42  & 87.88  & 84.10  & 78.46  & 71.13  \\
				RA\cite{randaugment}        & \textbf{91.74}  & 88.89  & 85.82  & 81.03  & 74.93  \\ \hline
				PDEN                          & 90.62 &  \textbf{88.91}& \textbf{87.03}& \textbf{83.71}&  \textbf{77.47}  \\ 
				\hline
			\end{tabular}
		}
	\end{center}
	\caption{Experiment results on CIFAR10-C dataset across different levels.}
	\label{exp_cifar10_level}
\end{table}

\begin{table*}[ht]
	\begin{center}
		\scalebox{0.90}{
			\begin{tabular}{llllllllllllll}
				\hline
				& \multicolumn{3}{c}{Weather} & \multicolumn{3}{c}{Blur} & \multicolumn{3}{c}{Noise} & \multicolumn{3}{c}{Digital} &    \\
				\cline{2-13}
				& Fog    & Snow    & Frost    & Zoom  & Defocus  & Glass & Speckle  & Shot & Impulse & Jpeg  & Pixelate  & Spatter & Avg. \\
				\hline
				ERM\cite{ERM}   & 65.92 & 74.36 & 61.57 & 59.97 & 53.71 & 49.44 & 41.31 & 35.41 & 25.65 & 69.90 & 41.07 & 75.36 & 56.15\\
				CCSA\cite{CCSA}  & 66.94 & 74.55 & 61.49 & 61.96 & 56.11 & 48.46 & 40.12 & 33.79 & 24.56 & 69.68 & 40.94 & 77.91 & 56.31\\
				d-SNE\cite{d-sne} & 65.99 & 75.46 & 62.25 & 58.47 & 53.71 & 50.48 & 45.30 & 39.93 & 27.95 & 70.20 & 38.46 & 73.40 & 56.96 \\
				GUD\cite{NIPS2018_7779}   & 68.29 & 76.75 & 69.94 & 62.95 & 56.41 & 53.45 & 38.45 & 36.87 & 22.26 & 74.22 & 53.34 & 80.27 & 58.26 \\
				MADA\cite{MADA}  & 69.36 & 80.59 & 76.66 & 68.04 & 61.18 & 61.59 & 60.88 & 60.58 & 45.18 & 77.14 & 52.25 & 80.62 & 65.59 \\
				AA\cite{autoaugment}    & 84.61 & 81.04 & 72.32 & 83.94 & 84.38 & 52.29 & 52.14 & 45.40 & 52.54 & 73.65 & 36.12 & 89.13 & 71.13 \\
				RA\cite{randaugment}    & \textbf{85.99} & 80.13 & 74.97 & \textbf{88.60} & \textbf{89.33} & 57.70 & 60.50 & 56.03 & 55.64 & 74.92 & 37.36 & \textbf{90.42} & 74.93 \\
				PDEN    & 69.64 & \textbf{81.81} & \textbf{84.50} & 83.73 & 82.15 & \textbf{60.13} & \textbf{79.31} & \textbf{81.28} & \textbf{66.79} & \textbf{85.24} & \textbf{70.82} & 79.38 & \textbf{77.47} \\
				\hline
			\end{tabular}
		}
	\end{center}
	\caption{The experimental result on CIFAR10-C. The model is trained on the clean data of CIFAR10 and evaluate on CIFAR10-C. We compared the accuracy of 19 types of corruption(only 12 corruptions are shown in the table) at level 5(the severest) in different methods.}
	\label{exp_cifar10_type}
\end{table*}

\begin{table*}[ht]
	\begin{center}
		\scalebox{0.93}{
			\begin{tabular}{lllllllllllll}
				\hline
				&        & \multicolumn{5}{l}{New York ish}            & \multicolumn{5}{l}{Old European Town}       &      \\ \cline{3-7} \cline{7-12}
				Source Domain                   & Method & Dawn & Fog & Night & Spring & Winter & Dawn & Fog & Night & Spring & Winter & Avg. \\ \hline
				\multirow{4}{*}{Highway/Dawn}   & ERM\cite{ERM}    & 27.80 & 2.73 & 0.93 & 6.80 & 1.65 & 52.78 & 31.37 & 15.86 & 33.78 & 13.35 & 18.70  \\
				& GUD\cite{NIPS2018_7779}    & 27.14 & 4.05 & 1.63 & 7.22 & 2.83 & 52.80 & 34.43 & 18.19 & 33.58 & 14.68 & 19.66   \\
				& MADA\cite{MADA}   &  29.10 & 4.43 & 4.75 & 14.13 & 4.97 & 54.28 & 36.04 & 23.19 & 37.53 & 14.87 & 22.33   \\
				& PDEN     &  \textbf{30.63} & \textbf{21.74} & \textbf{16.76}& \textbf{26.10} & \textbf{19.91} & \textbf{54.93} & \textbf{47.55} & \textbf{36.97} & \textbf{43.98} & \textbf{23.83} & \textbf{32.24} \\ \hline
				\multirow{4}{*}{Highway/Fog}    & ERM\cite{ERM}    & 17.24 & 34.80 & 12.36 & 26.38 & 11.81 & 33.73 & 55.03 & 26.19 & 41.74 & 12.32 & 27.16  \\
				& GUD\cite{NIPS2018_7779}    &  18.75 & \textbf{35.58} & 12.77 & 26.02 & 13.05 & 37.27 & \textbf{56.69} & 28.06 & 43.57 & 13.59 & 28.53   \\
				& MADA\cite{MADA}   & 21.74 & 32.00 & 9.74 & 26.40 & 13.28 & 42.79 & 56.60 & 31.79 & 42.77 & 12.85 & 29.00   \\
				& PDEN        &\textbf{25.61} & 35.16 & \textbf{17.05} & \textbf{32.45} & \textbf{21.03} & \textbf{45.67} & 54.91 & \textbf{37.38} & \textbf{48.29} & \textbf{20.80} & \textbf{33.83} \\ \hline
				\multirow{4}{*}{Highway/Spring} & ERM\cite{ERM}    & 26.75 & 26.41 & 18.22 & 32.89 & 24.60 & 51.72 & 51.85 & 35.65 & 54.00 & 28.13 & 35.02  \\
				& GUD\cite{NIPS2018_7779}     & 28.84 & \textbf{29.67} & 20.85 & 35.32 & 27.87 & 52.21 & \textbf{52.87} & 35.99 & 55.30 & 29.58 & 36.85  \\
				& MADA\cite{MADA}   & \textbf{29.70} & 31.03 & 22.22 & \textbf{38.19} & 28.29 & 53.57 & 51.83 & 38.98 & \textbf{55.63} & 25.29 & 37.47   \\
				& PDEN        & 28.17 & 27.67 & \textbf{27.53} & 34.30 & \textbf{28.85} & \textbf{53.75} & 51.53 & \textbf{46.87} & \textbf{55.63} & \textbf{30.61} & \textbf{38.49}  \\ \hline
			\end{tabular}
		}
	\end{center}
	\caption{Segmentation experiment results on SYNTHIA. We report the mIoU. All the models are trained on Highway and tested in New York ish and Old European Town.}
	\label{exp_synthia}
\end{table*}

\textbf{Comparison on SYNTHIA:} Follow the protocol in \cite{MADA}, we conducted three experiments, using Highway-Dawn, Highway-Fog and Highway-Spring as the source domain respectively, and taking all the weather in New York ish and Old European Town as the unseen target domains. The scene segmentation results(mIoU) are show in Tab \ref{exp_synthia}. Our approach improves the average mIoU compared to other approaches. When the source domain is highway-Dawn or Highway-fog, the improvement is greater.

\begin{figure}[t!]
	\centering
	\includegraphics[width=0.45\textwidth]{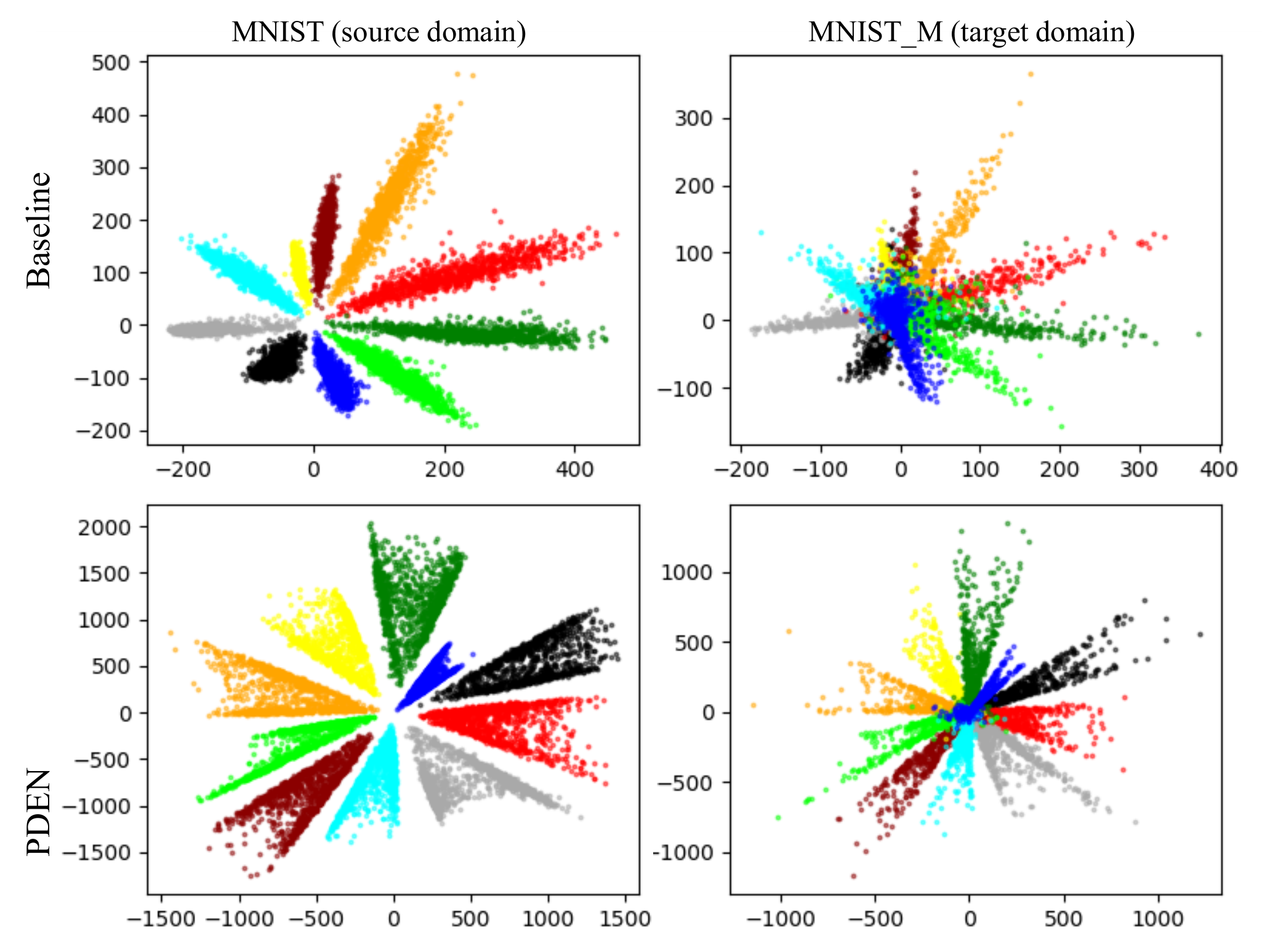}
	\caption{Visualization of different domains in the feature space. Rows 1 and 2 represent the feature space of baseline model and ours, respectively. Columns 1 and 2 represent the feature distribution of MNIST and MNIST\_M, respectively.}
	\label{vis_feature}
\end{figure}

\begin{figure}[h]
	\centering
	\subfigure[$K$]{
		\includegraphics[width=0.22\textwidth]{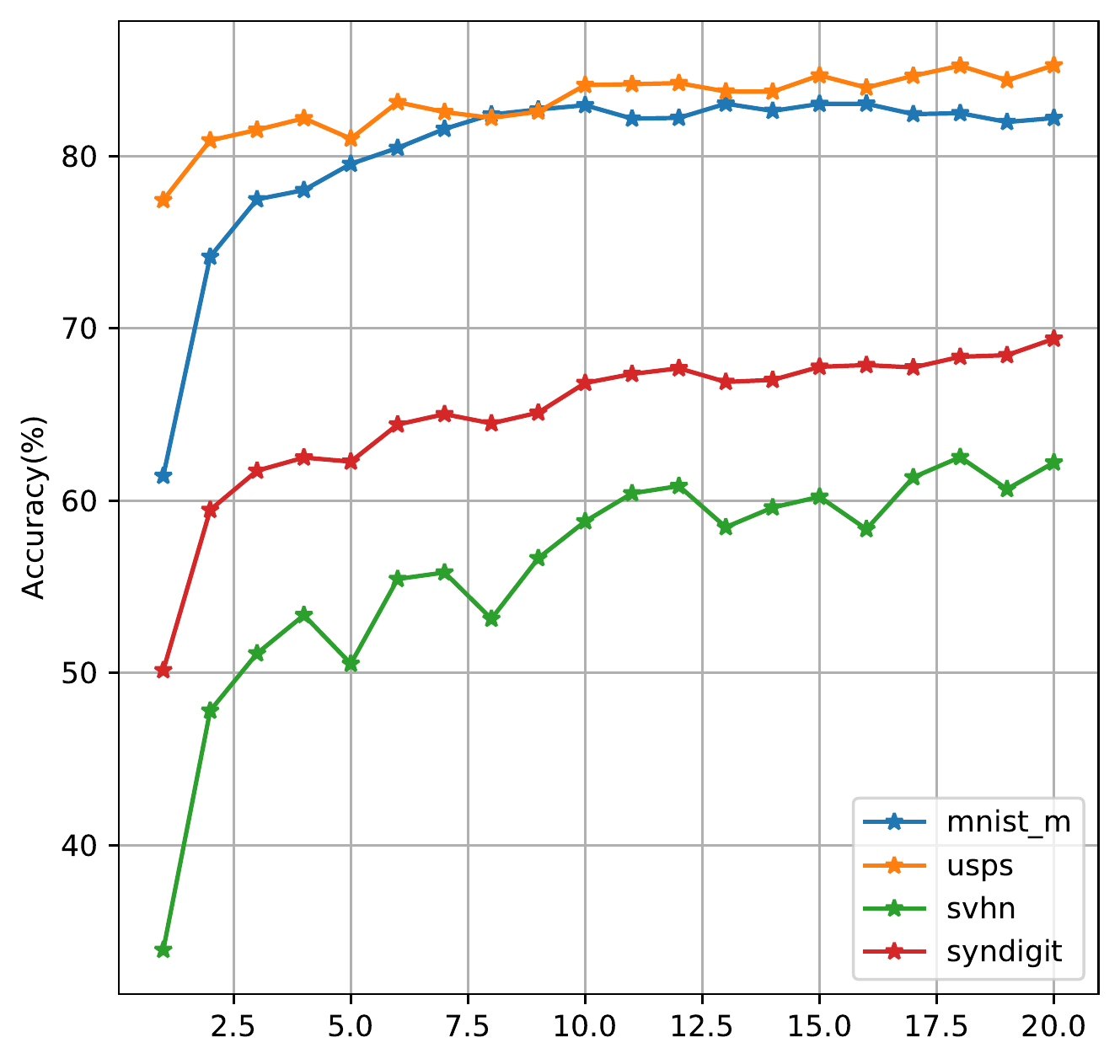}
		\label{ablation_k}
	}
	\subfigure[$w_{adv}$]{
		\includegraphics[width=0.22\textwidth]{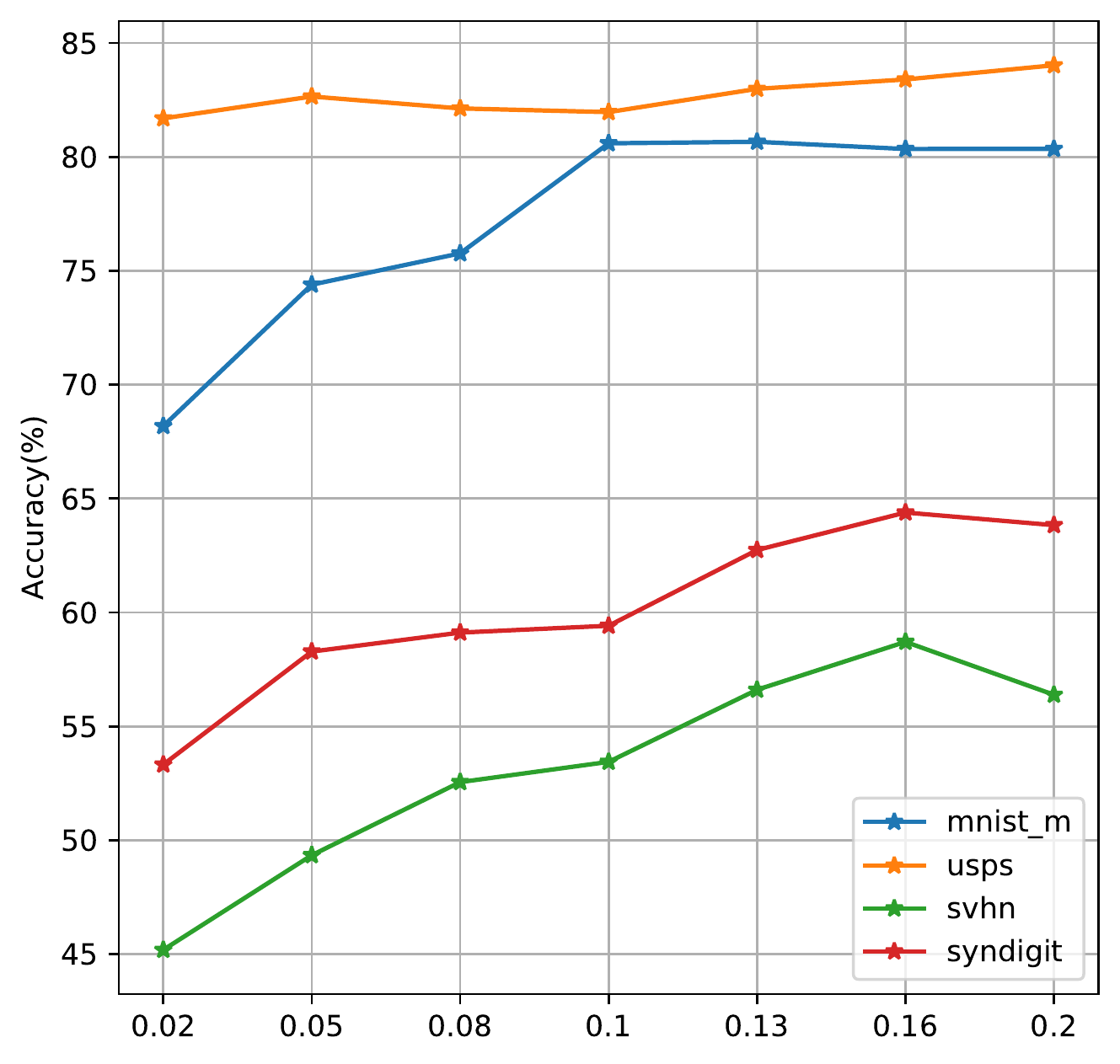}
		\label{ablation_winfo}
	}
	\subfigure[$w_{cyc}$]{
		\includegraphics[width=0.22\textwidth]{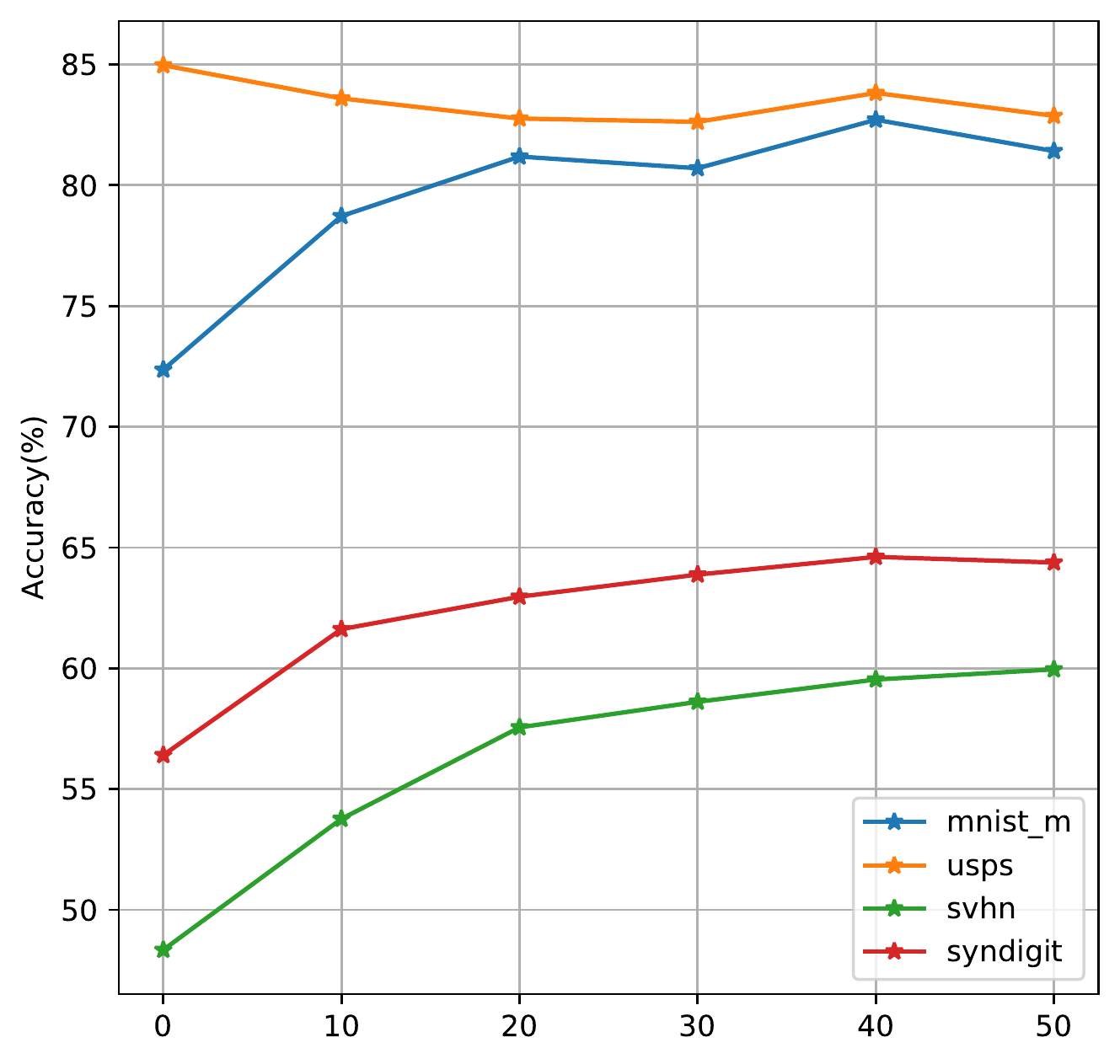}
		\label{ablation_wcyc}
	}
	\subfigure[$w_{div}$]{
		\includegraphics[width=0.22\textwidth]{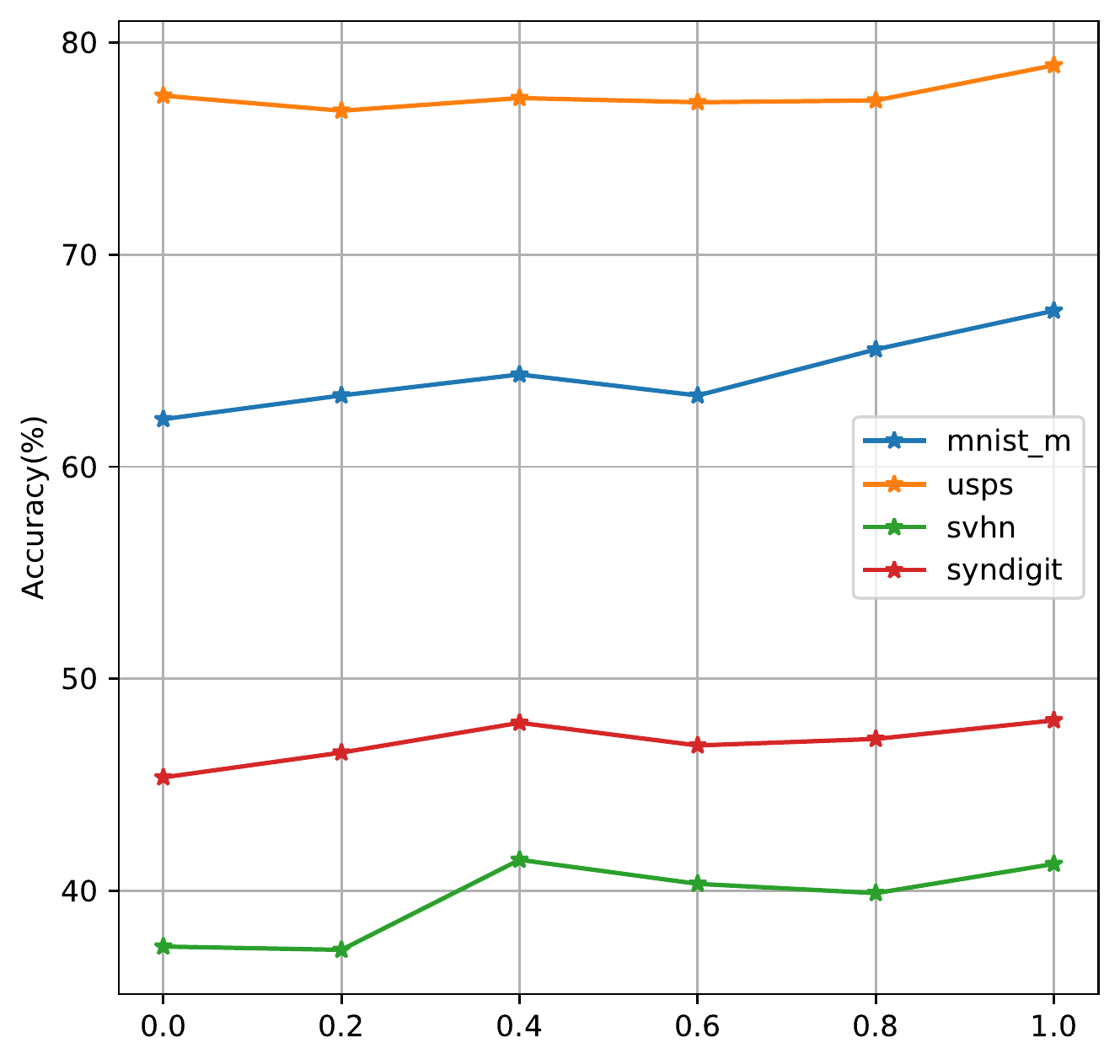}
		\label{ablation_wdiv}
	}
	\caption{Hyper-parameters tuning of $K, w_{adv}, w_{cyc}$ on the Digit dataset. }
	\label{Hyper-parameters}
\end{figure}

\subsection{Additional Analysis}

\textbf{Validation of $K$:} We study the effect of the hyper-parameters $K$ on the Digits dataset. We use the MNIST as the source domain, and take MNIST-M, SVHN, USPS and SYNDIGIT as the unseen target domains. The experimental result is shown in Fig.\ref{ablation_k}. We report the classification accuracy on the target domains when $K=1,2,...,20$. The accuracy is increased rapidly when $K$ is small, and gradually converges when $K$ is large. In experiments on Digits, we set $K=20$. In the Digits experiment in MADA\cite{MADA}, their approach performed best at $K=3$ and decreased as $K$ grew.This indicates that the domain generated by our approach is safer than MADA.

\textbf{Validation of $w_{adv}$:} We study the effect of the hyper-parameters $w_{adv}$ on Digits dataset. The experimental results are shown in Fig.\ref{ablation_winfo}. We report the classification accuracy on target domains when $w_{adv}=0.02, 0.05, 0.08, 0.1, 0.13, 0.16, 0.2$. We found that the accuracy increases with the increase of $w_{adv}$ on the unseen target domains. 

\textbf{Validation of $w_{cyc}$:} We study the effect of the hyper-parameters $w_{cyc}$ on Digits dataset. The experimental result is shown in Fig.\ref{ablation_wcyc}. We report the classification accuracy on MNIST-M, USPS, SVHN and SYNDIGIT when $w_{cyc}=0, 10, 20, 30, 40, 50$. On MNIST-M, SVHN and SYNDIGIT, the accuracy increases with the increases of $w_{cyc}$. On USPS, the classification accuracy did not change significantly (fluctuated within a small range) with the increase of $w_{cyc}$, mainly because of the high similarity between USPS and MNIST. 

\textbf{Validation of $w_{div}$:} We illustrate the effect of the hyper-parameters $w_{div}$ in Fig.\ref{ablation_wdiv}. For all the unseen domain in Digit dataset, the classification accuracy increases with the increases of $w_{dvi}$.

\textbf{Visualization of the feature space:} Fig.\ref{vis_feature} illustrates the difference in 2-d feature spaces between PDEN and the baseline models. For PDEN, the sample distribution of target domain is consistent with that of source domain. For the baseline model, most of the target samples are mixed in the feature space so that it is difficult to classify them.

\subsection{Evaluation of of Few-shot Domain Adaptation}
We also compared our methods in the experimental setting of few-shot domain adaptation \cite{motiian2017few}. In few-shot domain adaptation, data from the source domain $\mathcal{S}$ and a few samples from the target domain $\mathcal{T}$ are used to train the model. 

\begin{figure}[ht!]
	\centering
	\includegraphics[width=0.45\textwidth]{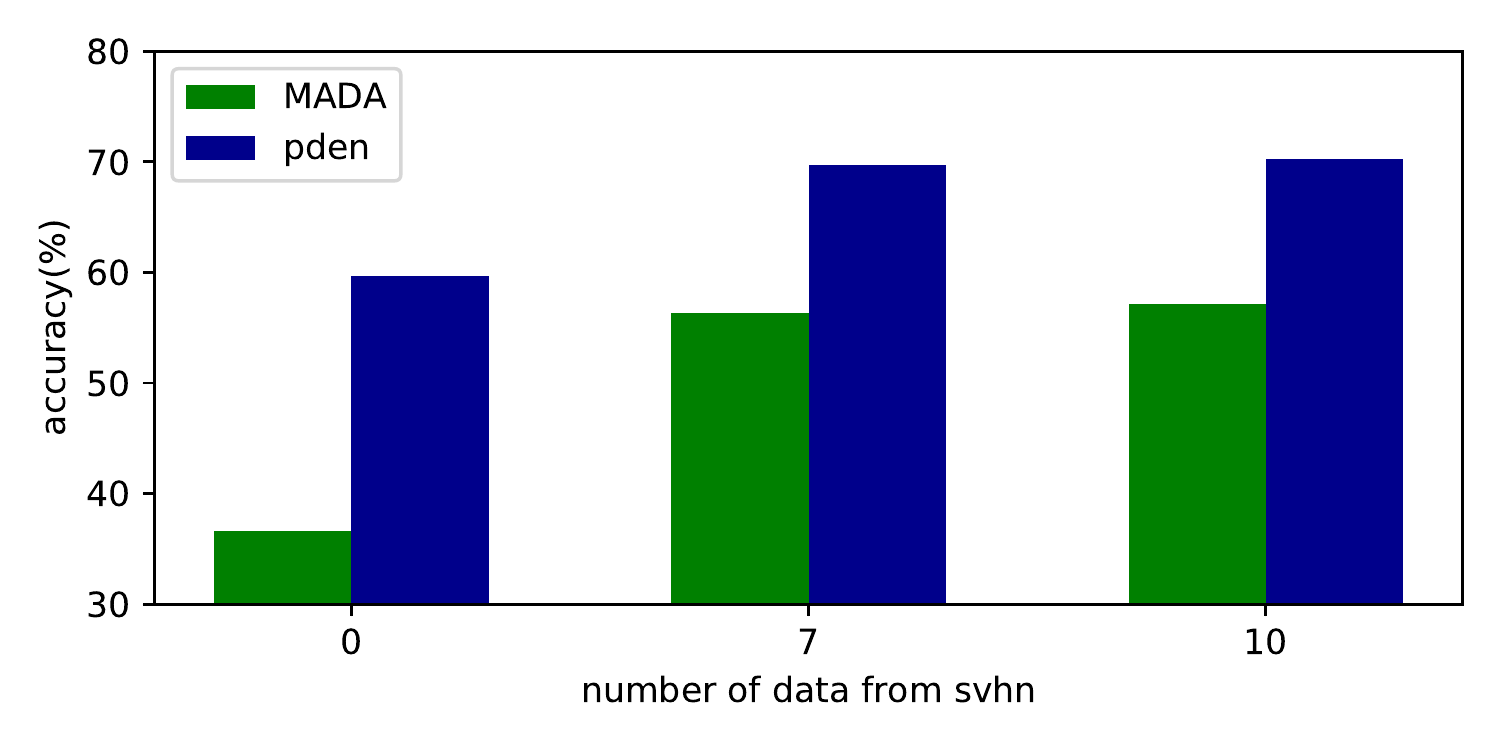}
	\caption{Few-shot domain adaptation experiment. We train the model with all the samples from MNIST and few samples from SVHN, and test the model on SVHN.}
	\label{few_shot}
\end{figure}

We use MNIST as the source domain and SVHN as the target domain. We first train the model on mnist with the proposed PDEN, and then finetune the model with few samples from SVHN. The model will be evaluated on SVHN, as shown in Fig. \ref{few_shot}. We found that finetuning with few samples from the target domain can significantly improve the performance of the model on the target domain. Compared with MADA, the proposed PDEN performs better in this case.

\section{Conclusion}
In this paper, we propose a single domain generalization learning framework to learn the domain-invariant feature, which can generalize the model to the unseen domains. We learn the generator to synthetic unseen domains, which share the same semantic information as the source domain. The domain-invariant representation can be learned by aligned the source and unseen domain distribution. We mine the hard unseen domains in which the domain-invariant representation cannot be extracted by the task model. The model will be more robust by adding these generated domains to the training set. The novel method PDEN proposed in this paper provides a promising direction to solve the single-domain generalization problem. 

\section*{Acknowledgements}
This work was supported by the National Key Research and Development Program of China (2018YFC0825202), and the National Natural Science Foundation of China (U1703261,61871004), and Beijing Municipal Natural Science Foundation Cooperation Beijing Education Committee: No. KZ 201810005002.

\clearpage
{\small
	\bibliographystyle{ieee_fullname}
	\bibliography{egbib}
}

\end{document}